\def\year{2017}\relax
\documentclass[letterpaper]{article}
\usepackage{aaai17}
\usepackage{times}
\usepackage{helvet}
\usepackage{courier}
\usepackage{amsmath}
\usepackage{amsfonts}
\usepackage{graphicx}
\usepackage{url}
\usepackage{pdfpages}
\usepackage{booktabs}

\usepackage{amssymb}  
\usepackage{caption}
\usepackage{subcaption}
\usepackage{pbox}
\usepackage{colortbl}

\begin{document}
%
\title{An ROS-based Shared Communication Middleware for \\Plug \& Play Modular Intelligent Design of Smart Systems }
\author{
Tathagata Chakraborti$^1\thanks{Most of the work was done during an internship (Summer, 2015) at the United Technologies Research Center, Berkeley, CA.}$~~~Siddharth Srivastava$^2$~~~Alessandro Pinto$^2$~~~Subbarao Kambhampati$^1$\\
$^1$Department of Computer Science, Arizona State University, Tempe, AZ 85281, USA\\ 
{\tt\small \{ tchakra2, rao \} @ asu.edu }\\
$^2$United Technologies Research Center, Berkeley, CA 94705, USA\\ 
{\tt \small \{ srivass, pintoa \} @ utrc.utc.com}
}

\nocopyright
\maketitle

\begin{abstract}
Centralized architectures for systems such as smart offices and homes are rapidly becoming obsolete due to inherent inflexibility in  their design and management. This is because such systems should not only be easily re-configurable with the addition of newer capabilities over time, but should also have the ability to adapt to multiple points of failure. Fully harnessing the capabilities of these massively integrated systems requires higher level reasoning engines that allow them to plan for and achieve diverse long-term goals, rather than being limited to a few predefined tasks. In this paper, we propose a set of properties that will accommodate such capabilities, and develop a general architecture for integrating automated planning components into smart systems. We show how the reasoning capabilities are embedded in the design and operation of the system, and demonstrate the same on a real-world implementation of a smart office.
\end{abstract}

\noindent The last decade has seen increasing interest in massively integrated systems that incorporate capabilities of sensing and actuation in order to adapt and respond to its environment dynamically. Popular examples of such systems include smart offices and smart homes and the closely related notion of Internet of Things. As these systems, collectively referred to as ``smart systems", become more and more advanced in their capabilities, and are able to assimilate increasingly complex components into their ecosystem, the major technological challenges faced are mainly in issues involving the integration of the diverse elements and functionalities, and providing them autonomy. 

\begin{figure}[tbp!]
\centering
\includegraphics[width=\columnwidth]{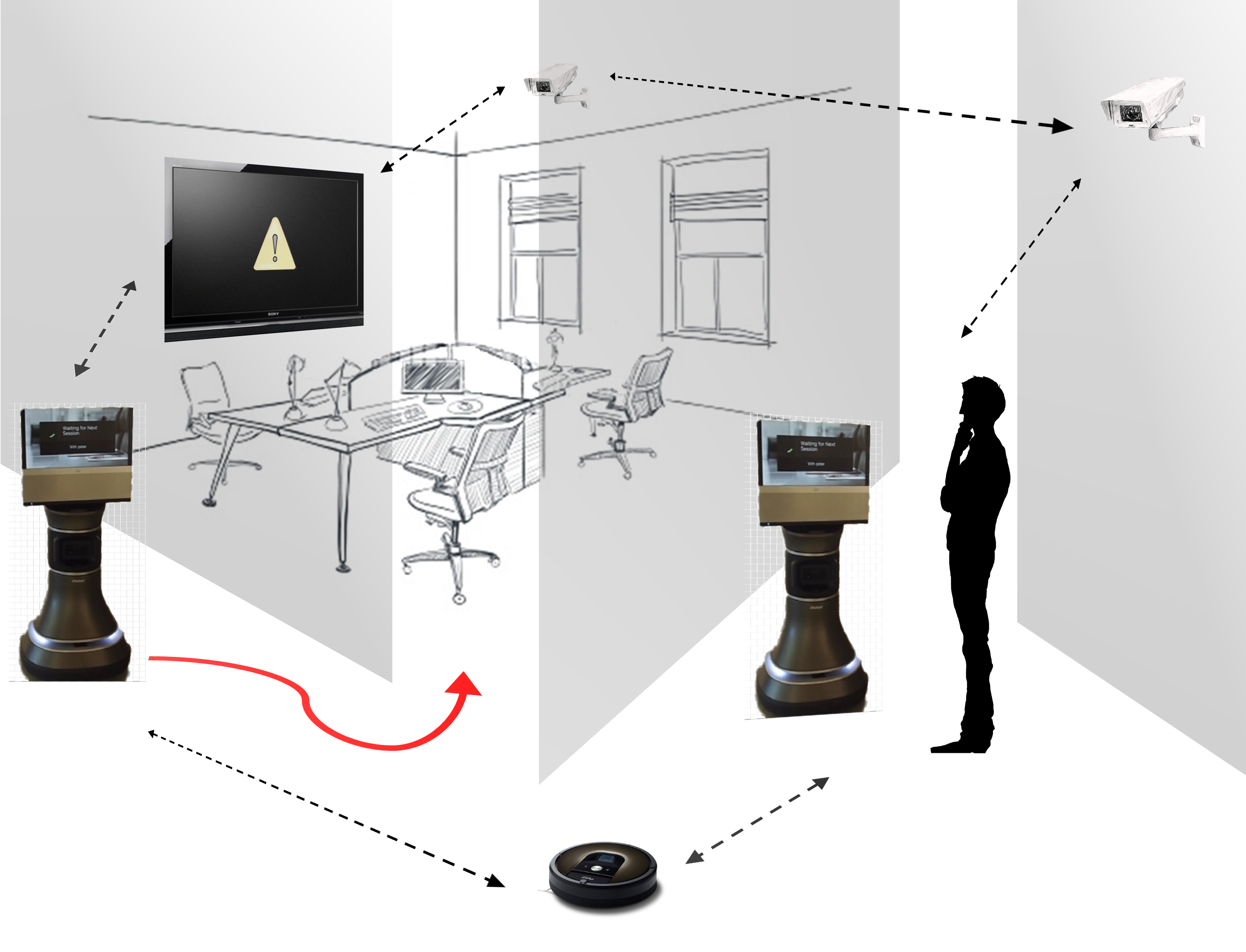}
\caption{Illustrative scenario in a smart office environment.} 
\label{scenario}
\end{figure}

Figure \ref{scenario} depicts an illustrative smart office environment. When the TV screen in the conference room is plugged in for the first time, the office should detect the addition of this new component, and complete the setup (resource allocation and integration) automatically, and understand that the conference room now has screen-casting abilities. This is referred to as {\bf plug \& play design}. Suppose now that the setup malfunctions and the screen is incapable of displaying anymore. The smart office may now deploy a video collaboration robot
to the conference room in order to resume the interrupted meeting. To do this the smart system must be equipped with the ability to sense the change in environment and {\bf reason at a high level} about the proper response to this change. We postulate that the building blocks of such systems should be in the form of {\bf modular, self-aware micro-agents} that can reason about their capabilities and the state of the environment, and achieve tasks independently or as a group. Further their design should be modular so as to be able to efficiently deal with the dynamic nature of the environment, as well as be abstracted to high level functionalities so as to handle the diversity of their internal implementations.

\subsubsection{Related Work} 
The integration of AI in smart rooms was first conceptualized in the MIT Intelligent Room Project \cite{617707}. Since then, the question of intelligent architecture design of ubiquitous systems has been explored extensively, most notably in Anand Ranganathan's seminal work \cite{Ranganathan2003} on \emph{context-aware} middleware design of pervasive systems. The adoption of automated planning technology in \texttt{Gaia} \cite{1301350} was certainly a first step towards augmenting high level reasoning capabilities in such environments. However, a recent study from \citeauthor{ieee} on the state-of-the-art in design of IoT environments identifies still persisting deficiencies in smart architectures, especially in terms of semantics of abstract APIs of system components and their generic high level reasoning capabilities. Indeed, while there has been significant interest in design of architectures for integrating task planning capabilities, especially in robotics applications, in recent times \cite{ICAPS1510619,mers}, there is still a large void to be filled when it comes to general purpose multi-agent embedded systems.

\subsubsection{Contributions} 

The main aim of this work is to outline a framework for integration of high level reasoning engines in a smart system in order to facilitate general distributed long-term decision making. To this end, we -

\begin{itemize}
\item[-] outline three principles - \emph{plug \& play, modular and intelligence} - for the design of smart systems.
\item[-] show how agents may follow protocols using abstract message classes to reason and communicate with each other, without accessing their internal implementations.
\item[-] create a prototype smart office model based on the discussed principles, and provide use cases and video demonstrations to illustrate these properties.
\end{itemize}
While some of these ideas have been investigated before, especially in the context of hardware, we will primarily focus on how the same principles may be applied in a unified manner at the higher level APIs as well.

\subsection*{Plug \& Play Modular Intelligent Agents} 
We argue that for smart systems to have the kind of functionalities we have discussed thus far, they should be designed to have three basic properties, as follows - 

\subsubsection*{Semantic Plug \& Play} This enables a system to -
\begin{itemize}
\item[-] Discover and integrate (i.e. resource allocation of) new components into the system automatically.
\item[-] Re-evaluate and update self-capabilities as the system components evolve with time.
\item[-] Deal with the dynamics of connection, e.g. individual components plugging in and out of the system due to network failures, internal failures, etc.
\end{itemize}
This concept of plug \& play is consistent with the traditional notion of plug \& play in computing, that facilitates the discovery and integration of hardware components (e.g. hotplug systems like USB and IEEE 1394 FireWire) usually during boot-time without requiring manual device configuration \cite{wiki-pnp}. 
At the same time we argue that the for a truly smart system, the traditional notion of plug \& play must extend to automatic dynamic re-configuration of hardware. We refer to this as \emph{semantic plug \& play}, wherein the system not only accommodates seamless hardware integration but also integration, at the API level, of abstract capabilities that may be reasoned with at a high level.
\subsubsection*{Intelligence} Finally, by intelligence, we mean that
\begin{itemize}
\item[-] Components of the system should be able reason about the current state of the environment, and the agents in it, in order to coordinate with other units in its ecosystem and achieve long-term tasks or goals. 
\item[-] The reasoning capabilities for the system should be realized in both centralized and decentralized fashion. This is critical as the system grows in complexity, in order to lower communication overhead and be suited to the limited computational power of embedded units, and also to avoid redundancy or backtracking during planning.
\end{itemize}%
\subsubsection*{Modular} A smart system is inherently distributed in nature - both physically and in terms of capabilities. Thus the architecture must itself be open and modular to support - 
\begin{itemize}
\item[-] Device agnosticism and, consequently, easy configuration across diverse platforms and hardware implementations.
\item[-] Handling of multiple points of failure, which is critical in such massively integrated systems, while also leveraging redundancies of fault-tolerant design.
\item[-] Agents capable of performing local computation. From the task planning standpoint, often only a small subset of the agents are required to complete a task. Further, the entire system capabilities is mostly not required by every single agent to compute its own contribution to the global plan. Thus making a centralized plan, or even planning individually with complete models, is often not necessary. 
\end{itemize}

\noindent The need for both centralized and decentralized reasoning engines is complimentary to a distributed approach to architecture design. Note, however, that the notion of a distributed architecture is distinct from whether the reasoning algorithm itself is distributed. Further, we will restrict ourselves to deterministic sequential planning as the primary reasoning capability, but the ideas discussed here can be easily extended to accommodate other forms of reasoning engines. 


\begin{figure}[t!]
\centering
\includegraphics[width=\columnwidth]{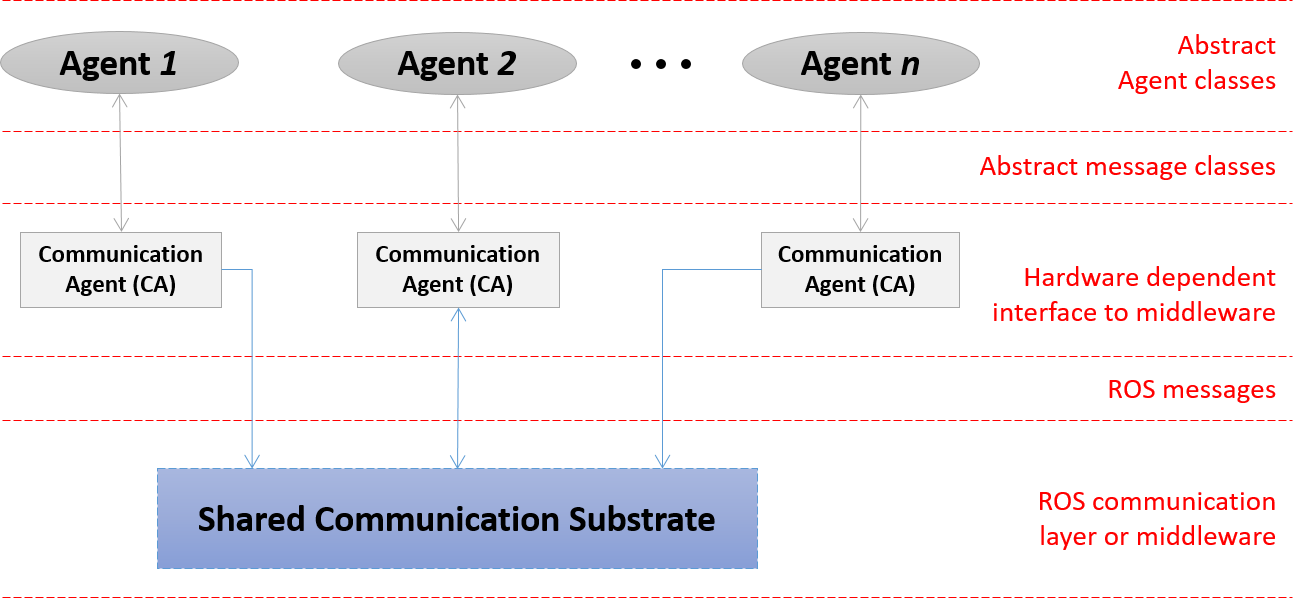}
\caption{General system architecture.}
\label{arch0}
\end{figure}

\section{Proposed Architecture Design}

Now we describe the general principles involved in our architecture design, that we will adopt in our smart system.

\subsubsection{Shared Communication Channel}
\label{arch}

The core of the architecture used is a single communication substrate as illustrated in Figure \ref{arch0}. All the individual components of the system or \emph{agents} are plugged into this shared communication layer, and they communicate between each other using abstract message classes. Both the description of the agents and the messages they use to communicate is at an abstracted level, that does not interfere with the actual implementation of the system's communication layer. Thus, the communication substrate can be swapped out and replaced with a different one, and the API's for the agents and message classes will remain exactly the same. Each agent, however, has a Communication Agent (CA) built into it that acts as the interface to the communication layer. Thus the communication agent itself is implementation specific, and needs to be reconfigured according to the environment.

\subsubsection{Communication Protocols}
The abstract message classes provide a common language for the agents to communicate with. This closely follows the specifications of agent communication language (ACL) laid out by the IEEE Foundation for Intelligent Physical Agents (FIPA) in \cite{fipa}. This language is defined by 11 \emph{performative types} that perform their specific roles during an information exchange. The types used in the paper and their typical roles are described in Table \ref{table1}.

\begin{table}[t!]
\small
\centering
\caption{Abstract message classes - types and usage.}
\label{table1}
\begin{tabular}{@{}l|l@{}}
\toprule
\multicolumn{1}{l|}{\textbf{performative type}} & \multicolumn{1}{l}{\textbf{typical functions}}                                                                                                                                                                    \\ \midrule
agree                                 & - reply with plan proposal (from actuator)                                                                                                                                                               \\
\arrayrulecolor{gray}\hline
cancel                                & - cancel existing commitment                                                                                                                                                                             \\
\arrayrulecolor{gray}\hline
refuse                                & - refuse deployed request                                                                                                                                                                                \\
\arrayrulecolor{gray}\hline
request                               & \begin{tabular}[c]{@{}l@{}}- request plan from Plan Requester\\ - request plan from Planner\\ - request execution of plan\\ - request Actuator to execute\\ - request for state/goal update\end{tabular} \\
\arrayrulecolor{gray}\hline
call for proposals                    & - broadcast call for plan proposals                                                                                                                                                                      \\
\arrayrulecolor{gray}\hline
propose                               & - reply with plan proposal (from planner)                                                                                                                                                                \\
\arrayrulecolor{gray}\hline
accept                                & - accept proposal                                                                                                                                                                                        \\
\arrayrulecolor{gray}\hline
reject                                & - reject proposal                                                                                                                                                                                        \\
\arrayrulecolor{gray}\hline
inform                                & \begin{tabular}[c]{@{}l@{}}- inform action model on activation\\ - update self state\end{tabular}                                                                                                        \\
\arrayrulecolor{gray}\hline
query                                 & - ask for information from the UI                                                                                                                                                                        \\
\arrayrulecolor{gray}\hline
confirm                               & \begin{tabular}[c]{@{}l@{}}- monitor action / plan execution\\ - relay observations
\end{tabular}\\                                                                       \midrule
\end{tabular}
\end{table}


We use ROS \cite{ros} as our communication substrate. All agents are plugged into a central ROS server, and they communicate with each other using \emph{ROS messages}. The abstract message API thus acts as a wrapper for the lower level ROS messages. The Communication Agent implements a ROS node that publishes and subscribes to the appropriate topics. Each of the message types have their specific global ROS topic that all agents have access to, where messages of corresponding type are published.
Though the roles of the message types are specified as part of the language description, the semantics of a message of particular type is decided by the agent that processes it. 


\subsubsection{Centralized Planning}

\begin{figure}
    \centering
    \begin{subfigure}[b]{\columnwidth}
    \includegraphics[width=\columnwidth]{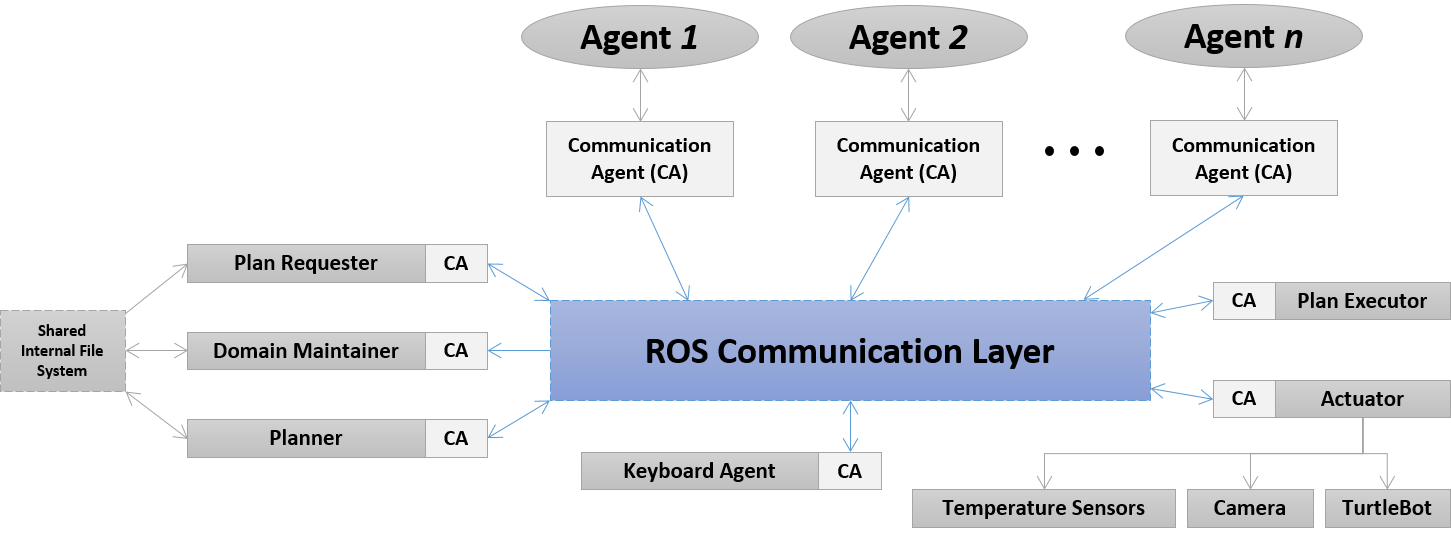}
\caption{Centralized system design - agents have access to a central planning engine via the communication substrate.}
           \label{fig3:a}
    \end{subfigure}
    \begin{subfigure}[b]{\columnwidth}
\vspace{10pt}
\includegraphics[width=\columnwidth]{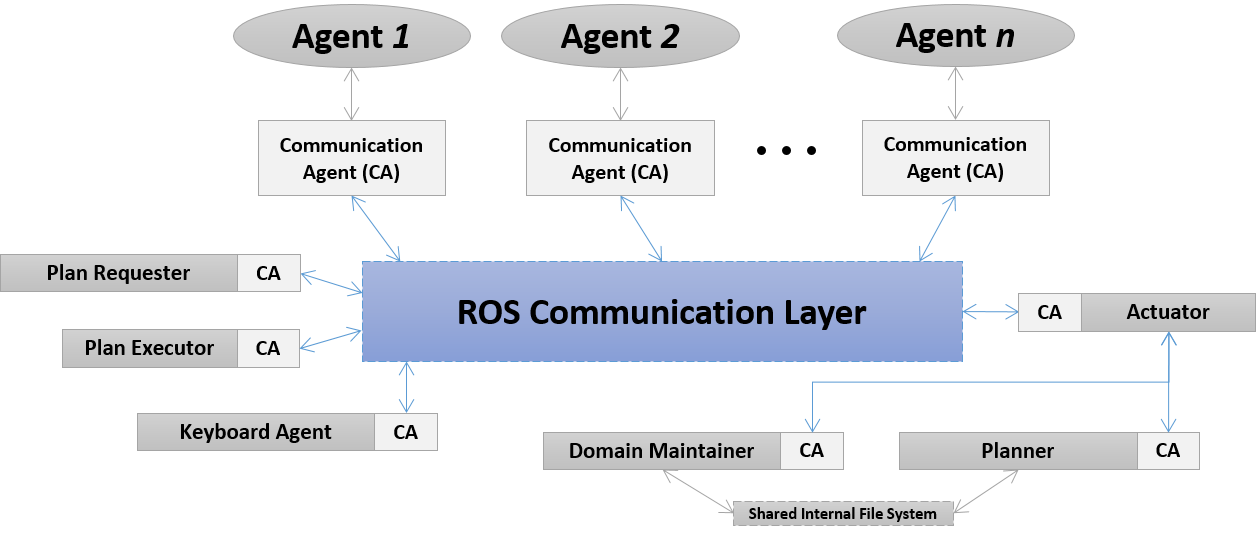}
\caption{Decentralized system design - now the actuators have their own planners and domain maintainers.}
           \label{fig3:b}
    \end{subfigure}
    \caption{System architectures adopted from Figure \ref{arch0}.}
\label{fig3}
\end{figure}

The architecture for centralized planning is shown in Figure \ref{fig3:a}. 
The agents plugged into the shared communication substrate are of three basic types - some of them handle the user interface (keyboard), some are capable of interacting with the environment (different types of actuators like the camera, the mobile base and stationary sensors) while the rest of them deal with the reasoning tasks of the system.
This framework supports centralized planning since global agents deal with specific tasks of the planning process, and all agents report and hear back from these.

\subsubsection{Decentralized Planning}

The architecture in Figure \ref{fig3:b} is adopted to facilitate decentralized planning capabilities of the system. Here actuators (i.e. any agent that can influence the environment) are equipped with their own Planners and Domain Maintainers. Without going into details of these individual agents, this means that actuators are now capable of building their own plans. The processing of the plan request and execution is however still done in a centralized fashion so that proposals from individual agents can be evaluated. 

\subsection{Agent Ecosystem}
\label{agents}

\begin{figure}[tbp!]
  \begin{center}
    \includegraphics[width=\columnwidth]{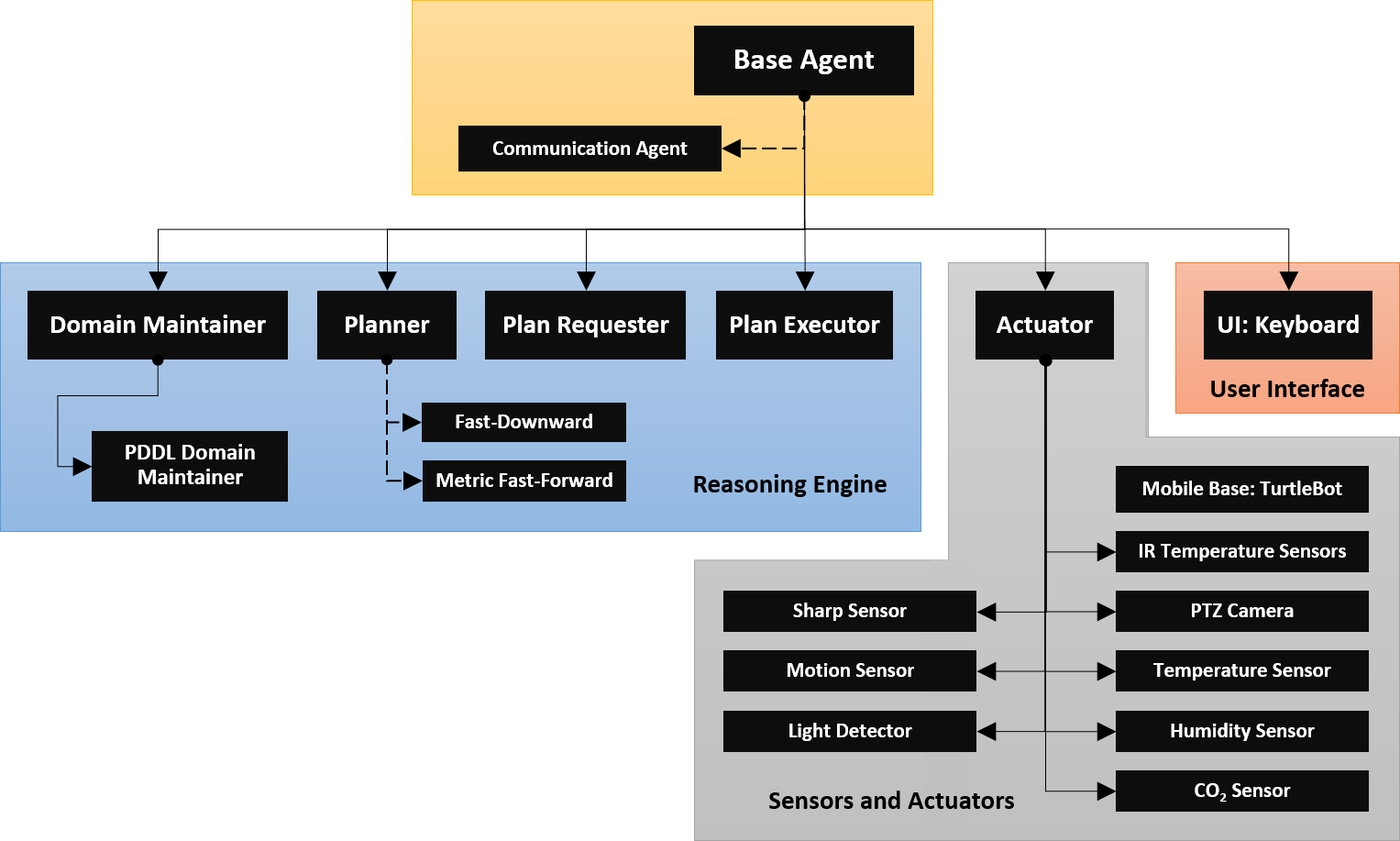}
  \end{center}
\caption{Agent class hierarchies.} 
\label{fig5}
\end{figure}

We will now describe the architecture illustrated in Figure \ref{fig3} and the hierarchy of implemented agent classes in Figure \ref{fig5}. 

\subsubsection{The Base Agent}

is the basic agent class that all other agents inherit from. It has abilities to send/receive messages with the help of the communication middleware attached to it, and can perform basic processing on these messages. 

\subsubsection{\textnormal{\emph{The Communication Agent (CA)}}} is the interface between the Base Agent API and the communication substrate. The primary functionality of the CA is thus to assist in the relaying of messages back and forth between agents through the shared communication channel. The CA also has special callback functions that alert its parent agent about specific messages addressed to it. The CA can also perform secondary tasks such as listening to the communication channel for agent \emph{heartbeats}, perform queue operations on agent death/activation or in response to messages processing, cache broadcast messages for processing, and so on.

\subsubsection{The Reasoning Engine} 
consists of agents concerned with task planning and execution, with the help of which the system can reason with models of its components at a high level without accessing their internal implementation. In the distributed version, the agents can continue using the same components to compute plans locally - only the architecture and communication protocols change. Four specific agents make up this reasoning engine -

\subsubsection{\textnormal{\emph{Plan Requester}}} sets up a request for plan computation (upon request from either the user or some other agent) and calls the Planning Agent for a plan. The plan request compiles all the state and domain information that may be required to solve a planning problem.

\subsubsection{\textnormal{\emph{Planning Agent}}} computes the actual plan (and replies back with the plan and plan status), given a plan request from the Plan Requester with specific domain and problem descriptions. Currently it is integrated with the state-of-the-art planners \texttt{FAST-DOWNWARD} \cite{fd1} and \texttt{Metric FAST-FORWARD} \cite{ff}. 

\subsubsection{\textnormal{\emph{Plan Executor}}} is tasked with interfacing with Actuators. It receives a request for a plan to be executed, transforms these higher level actions into lower level methods that the specific actuators have access to, and sends specific action requests to the concerned agents. It is also responsible for plan monitoring and error handling during execution.

\subsubsection{\textnormal{\emph{Domain Maintainer}}} is tasked with maintaining the evolving capabilities of the system. These may be both global or local capabilities according to whether the centralized or distributed architecture is being used. It can inform the reasoning engine about changes to the current state, goal/cost function, and available actions and observations. This is the agent that is primarily responsible for managing the plug \& play mechanism of the system by listening to the \emph{heartbeat} of agents and re-evaluating the system capabilities dynamically.
The agent models are currently represented in \texttt{PDDL} \cite{pddl} - \texttt{PDDL} support is the same as that of the underlying planning software being used.

\subsubsection{Actuators} inherit the Base Agent functionalities and add capabilities for interaction with the environment (e.g. sensing and actuation). They can also announce their self-capabilities to the system (by broadcasting their action models at regular intervals - this is referred to as its \emph{heartbeat} and is initiated when it first becomes active) and execute actions within advertised capability when requested. Specialized actuators inherit from the basic Actuator class -

\subsubsection{\textnormal{\emph{TurtleBot}}} is a mobile base - it can move between (pre-defined) poses in the map. It can also carry non-stationary sensors. The onboard laptop also contributes to the distributed keyboard interface to the users.

\subsubsection{\textnormal{\emph{PTZ Camera}}} can move (PTZ), record video, snap pictures, and detect motion \cite{motion}. Its controls are also accessible through the keyboard interface.

\subsubsection{\textnormal{\emph{Temperature / Humidity Sensor}}} reports temperature at its current location (stationary).

\subsubsection{\textnormal{\emph{Temperature IR Sensor}}} reports temperature at its current location (variable, when mounted on a mobile base).

\subsubsection{\textnormal{\emph{Other Sensors}}} The platform supports a wide variety of sensors, including CO$_2$ Sensor, Motion Sensor, IR Distance Adapter with Sharp Sensor, Wide Range Light Sensor, IR Reflective Sensor, etc.
%
%
All these sensors are stationary and need to connect to a SBC. We also have integration capabilities directly to the computer with a mini-USB port through a SBC Interface Kit (however, this requires power). 

\subsubsection{User Interface}

The user interface is through the keyboard (and maintained by the Keyboard Agent). The keyboard offers a simple platform for users to request plans or actions from the system. It also provides simple tools for debugging. 
One additional advantage of using the keyboard is that it can be conveniently distributed across the system through multiple shared screens. Currently the central server and also the laptops on the mobile bases shares the keyboard interface (e.g. through \texttt{tmux}).
In terms of specific capabilities, through the Keyboard Agent the user can (1) request for plans/actions; (2) modify state or goal descriptions; and (3) respond to information requests from agents.

\subsection{Flow of Control in Centralized Plan Execution}

We will now illustrate an example of flow of control within the agents for centralized planning (Figure \ref{fig:flow:c1}). The setting consists of stationary temperature sensors at various locations, and mobile bases with on-board sensors.

\begin{itemize}
\item[-] Initially all sensors are active and broadcasting their action models. The Domain Maintainer compiles these into a composite high level model of the system.
\item[-] The User requests a requirement (e.g. report temperature from all locations) through the Keyboard Agent, which it forwards on to the Plan Requester.
\item[-] The Plan Requester receives the request, sets up the problem/domain by converting the requirement to a high level PDDL goal, and requests a plan from the Planner.
\item[-] The Planner computes a plan (which involves every sensor reporting their measurements from their corresponding locations) and responds back to the Plan Requester, which forwards it to the Executor.
\item[-] The Plan Executor goes through the actions in the plan sequentially, translates them into lower level methods of the agents involved, and dispatches these action requests to the appropriate actuators (it requests every stationary temperature node to respond with their measurements). 
\item[-] The Executor receives confirmation of execution success from the actuator and moves on to the next action in the plan. 
The plan execution is complete when the Executor has received acknowledgments from all the sensors.
\item[-] Now some of the stationary sensors become inactive. The Domain Maintainer detects the missing heartbeats and updates the system capabilities to reflect this.
A new plan request is formed, and the above procedure ensues again, until finally, the Executor dispatches mobile bases to compensate for the loss of stationary sensors.
\item[-] The way the mobile bases are allocated tasks depends on action costs and map of the environment (which is reflected in the problem and domain descriptions compiled by the Domain Maintainer).
\end{itemize}


\subsection{Flow of Control in Decentralized Plan Execution}

\begin{figure}[!ht]
    \centering
    \begin{subfigure}[b]{\columnwidth}
    \includegraphics[width=\columnwidth]{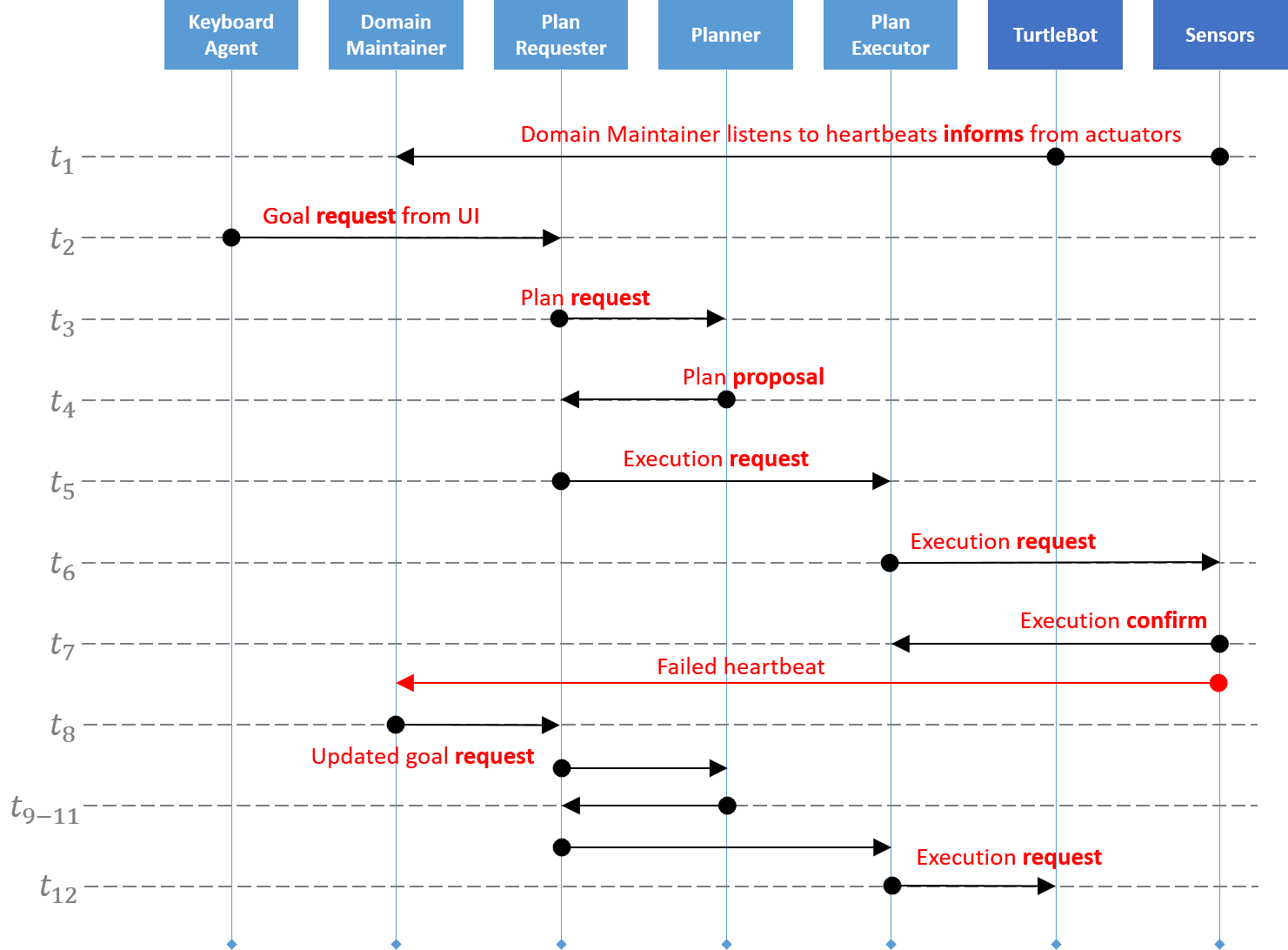}
	\caption{Flow of control in centralized plan execution.}
           \label{fig:flow:c1}
    \end{subfigure}\vspace{20pt}
    \begin{subfigure}[b]{0.9\columnwidth}
    \includegraphics[width=\columnwidth]{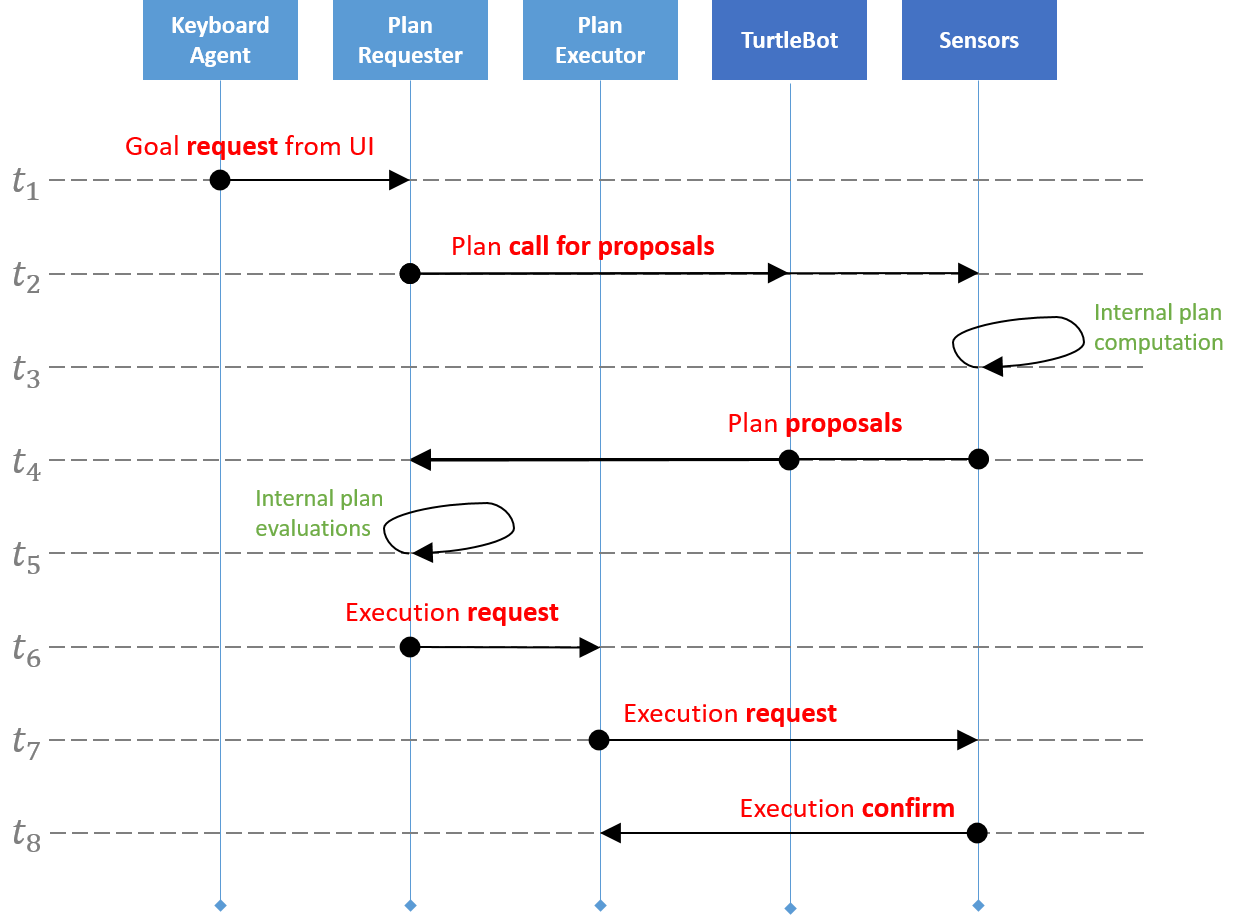}
	\caption{Flow of control in decentralized plan execution.}
           \label{fig:flow:c2}
    \end{subfigure}
    \caption{Interaction Diagrams.}
\label{fig:case:c}
\end{figure}

The flow of control in the decentralized scenario is slightly different since their is no single global Planner and Domain Maintainer. This means that the Plan Requester has to adopt an interactive procedure with the agents to come up with a global solution, as illustrated in Figure \ref{fig:flow:c2}.

\begin{itemize}
\item[-] As before, the user requests for a goal through the Keyboard, which forwards the request to the Plan Requester.
\item[-] The Requester now sends out a call for proposal to every agent plugged into the system by means of a broadcast.
\item[-] Each agent receives the request for a plan and computes a plan to address the goal(s) mentioned in the request. The agent does this by invoking its internal planner which has access to local domain knowledge as compiled by its personal Domain Maintainer.
\item[-] The agents who are able to contribute a plan respond to the call for proposal with the proposed plan.
\item[-] The Plan Requester listens to the communication channel and accumulates the plan proposals it receives. After the wait time is over, it goes through the cached responses and selects the cheapest candidate for execution.
\item[-] The Requester dispatches this plan to the Plan Executor for execution, and the rest of the control flows as described previously in case of the centralized system.
\end{itemize}

\subsection{Demonstrations\footnote{Video recordings of these demonstrations could not be made available for public release.}}
\label{demo}

\begin{figure}[t!]
  \begin{center}
    \includegraphics[width=0.9\columnwidth]{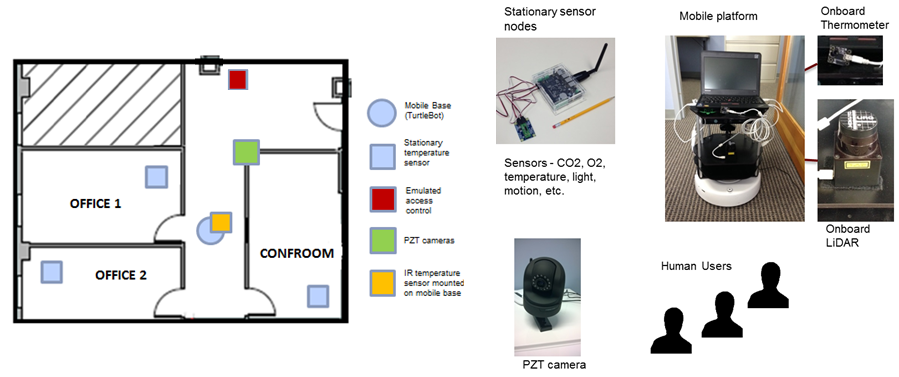}
  \end{center}
\caption{Test bed - a smart office environment.}
\label{fig2} 
\end{figure}

Our smart environment, shown in Figure \ref{fig2}, consists of two office rooms and one conference room (each equipped with a stationary temperature sensor) connected by a single corridor (that has a PZT camera equipped with motion detection). The corridor is patrolled by a mobile base, which is a TurtleBot that also carries an IR temperature sensor. The positions of these agents are shown in the figure. Besides, there are human users that can also interact with the system. 

\subsubsection{Scenario 1}
The goal here is to collect temperature measurements from all rooms. \texttt{Office2} and \texttt{confroom} are equipped with stationary sensors but \texttt{office1} is not - so the system dispatches the TurtleBot to report temperature from \texttt{office1}. This shows how the system can reason with the state of the environment and higher level models of its components to achieve goals. When the temperature sensor in \texttt{office2} goes down, the system detects this and re-evaluates its capabilities automatically and dispatches the TurtleBot to \texttt{office2} instead. This is an instance where the plug \& play architecture comes into use, as the system is able to adapt and re-plan accordingly. 

\subsubsection{Scenario 2} 
Now the sensor in \texttt{confroom} goes down, and the system adapts and replans accordingly. This again shows how the system responds not only to evolving capabilities but also to changing environment and (maintenance) goals. 

\subsubsection{Scenario 3} 
We now emulate a simple access control mechanism, where the camera detects movement at the entry and requests for the person entering to be validated. The system dispatches the TurtleBot to the location so that the person can log in through the on-board screen. Here the agents themselves respond to the changing environment and request for a plan update. This is also an instance where the user gets to interact with the system through the distributed interface.

\subsection{Future Work}

We end by envisioning a few extensions to the proposed architecture and motivating some future avenues of research.

\subsubsection{Distributed Communication Substrate}

\begin{figure}[t!]
  \begin{center}
    \includegraphics[width=\columnwidth]{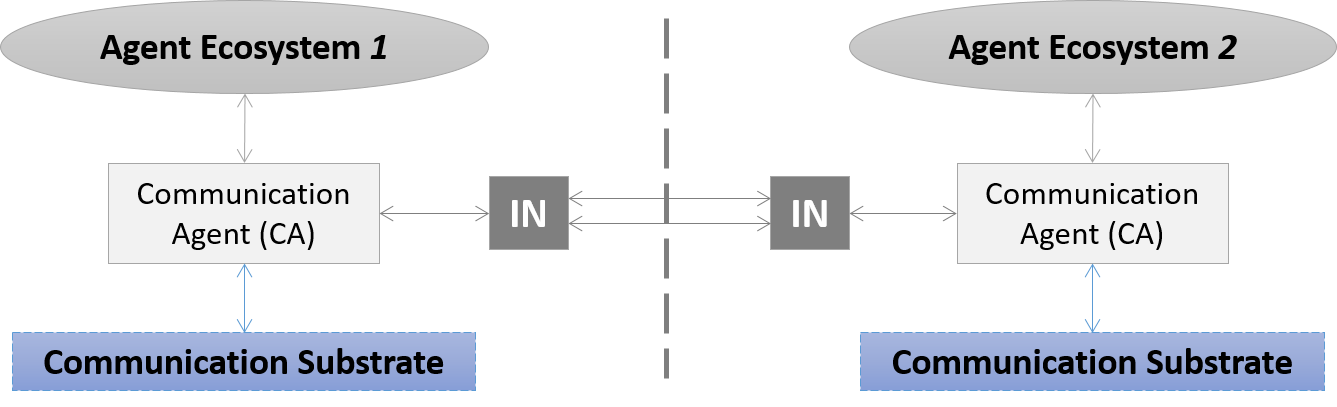}
  \end{center}
\caption{Separate ROS masters for each agent ecosystem, with the Interfacing Node (IN) acting as the mediator for common channels of communication.}
\label{fig6}
\end{figure}

As the number of agents increase, the communication load on a single ROS master will eventually become prohibitive. One way to alleviate concerns of scalability is to adopt different ROS masters for each \emph{agent ecosystem} \ref{fig6}, where groups of agents are using their own ROS environment, that by itself works as described in this paper; so that these separate ecosystems interact with each other through Interfacing Nodes (IN) to share messages of mutual interest. This architecture will not only deconflict identical ROS topics across ecosystems, but also preclude the need for ROS namespace mappings, and privatize topics that does not involve other agents.

\subsubsection{Distributed Planning}

The protocols for decentralized planning discussed in this paper will only work, of course, for non-disjunctive goals and when agents will not require mutual cooperation. However, the same protocols can support sharing of other information too, such as domain models, state information, etc. We can thus equip the Plan Requester with more sophisticated plan processing algorithms, like auctioning and merging \cite{Borrajo:2013:MPP:2484920.2485111}, to come up with a full-fledged distributed planning protocol. Given the framework can already support this, this also opens pathways to leverage existing literature in planning by distributed search. \cite{cont2}.



\bibliographystyle{aaai}
\bibliography{bib}

\end{document}